\pdfoutput=1

\documentclass[11pt]{article}

\usepackage[final]{acl}

\usepackage{times}
\usepackage{latexsym}
\usepackage{amsmath}
\usepackage{multirow}
\usepackage{booktabs}
\usepackage{amssymb}
\usepackage{bbding}
\usepackage{pifont}
\usepackage{utfsym}

\usepackage[T1]{fontenc}

\usepackage[utf8]{inputenc}

\usepackage{microtype}

\usepackage{inconsolata}

\usepackage{graphicx}

\newcommand*\samethanks[1][\value{footnote}]{\footnotemark[#1]}

\title{ARGUS: Policy-Adaptive Ad Governance via Evolving Reinforcement with Adversarial Umpiring}

\author{Deyi Ji$^1$ ~~Junyu Lu$^{2}$\thanks{Junyu Lu and Xuanyi Liu participated in this work while interning at Tencent as part of the Tencent Rhino-Bird Research Elite Program, with Deyi Ji as the program leader.} ~~Xuanyi Liu$^3$\samethanks ~~Liqun Liu$^1$ ~~Hailong Zhang$^1$ ~~Peng Shu$^1$ \\ ~~\textbf{Huan Yu}$^1$ ~~\textbf{Jie Jiang}$^1$ ~~\textbf{Tianru Chen}$^4$ ~~\textbf{Lanyun Zhu}$^5$\thanks{Corresponding Author.} \\
  $^1$Tencent ~~~ $^2$Dalian University of Technology ~~~ $^3$Peking University   \\ $^4$Zhejiang University  ~~~ 
$^5$Tongji University  \\
  \texttt{deyiji@tencent.com~ dutljy@mail.dlut.edu.cn~ xuanyi@stu.pku.edu.cn} \\
  \texttt{\{liqunliu,lericzhang,archershu,huanyu,zeus\}@tencent.com}  \\ \texttt{tianrun.chen@zju.edu.cn~ zhulanyun1999@gmail.com} 
  }

\begin{document}
\maketitle

\begin{abstract}
Online advertising governance faces significant challenges due to the non-stationary nature of regulatory policies, where emerging mandates (e.g., restrictions on education or aesthetic anxiety) create severe label inconsistencies and reasoning ambiguities in historical datasets. In this paper, we propose ARGUS, a policy-adaptive governance system that enables evolving reinforcement through multi-agent adversarial umpiring. ARGUS addresses the sparsity of new policy data by employing a three-stage framework: (1) Policy Seeding for initial perception; (2) Adversarial Label Rectification, which utilizes a ``Prosecutor-Defender-Umpire'' architecture to resolve conflicts between stale labels and new mandates; and (3) Latent Knowledge Discovery, which employs a tripartite dialectical discussion to unearth sophisticated, ``gray-area'' violations. By leveraging RAG-enhanced policy knowledge and Chain-of-Thought synthesis as dynamic rewards for reinforcement learning, ARGUS synchronizes its reasoning pathways with evolving regulations. Extensive experiments on both industrial and public datasets demonstrate that ARGUS significantly outperforms traditional fine-tuning baselines, achieving superior policy-adaptive learning with minimal gold data.

\end{abstract}

\begin{table*}[htbp]
\centering

\label{tab:case_study_analysis}
\small
\begin{tabular}{p{0.18\linewidth} | p{0.15\linewidth} | p{0.6\linewidth}}
\toprule
\textbf{Case Image} & \textbf{Agent Role} & \textbf{Reasoning Process} \\ \midrule
\multirow{4}{*}{\begin{minipage}{.18\textwidth} \centering \includegraphics[width=0.8\linewidth]{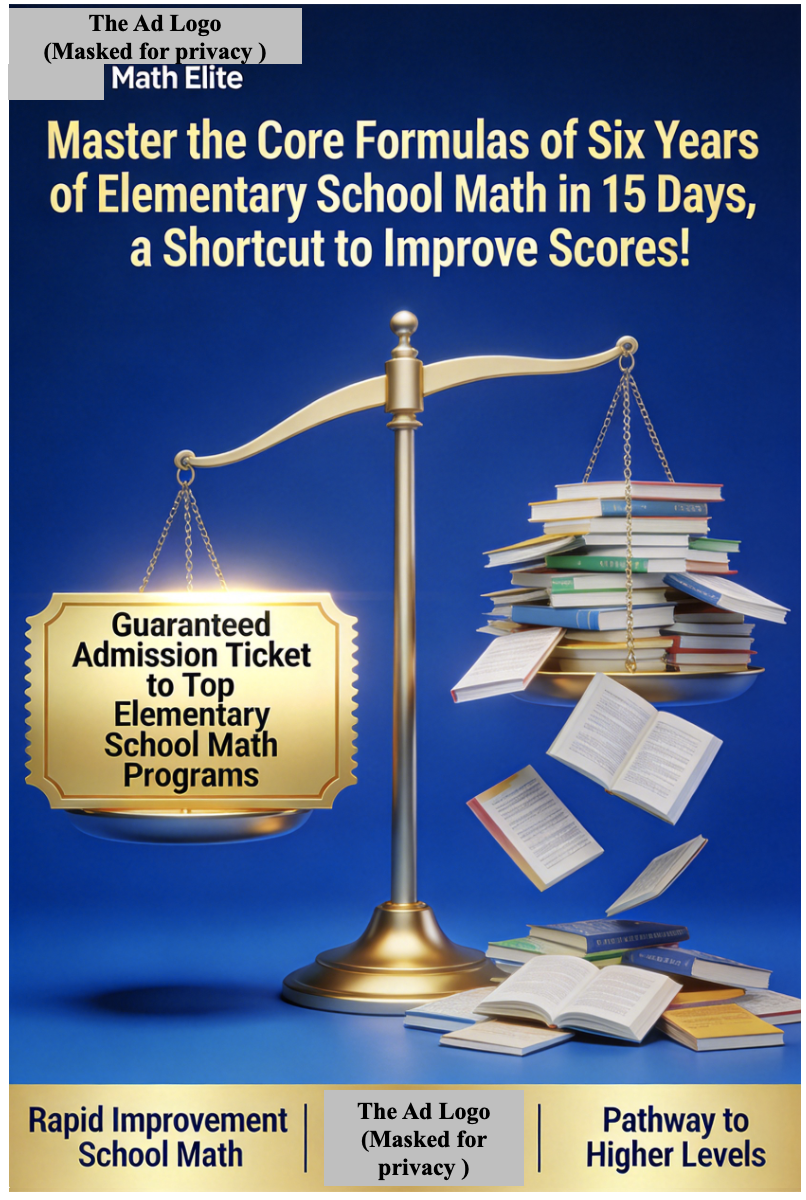} \\ \textbf{Case 1: K12-T} \end{minipage}} 
& \textbf{Prosecutor} & The ad promises to ``Master six years of math in 15 days'', which is a typical false claim violating educational laws. The ``Admission Ticket'' metaphor implies a shortcut to elite schools, exploiting parental anxiety for enrollment. \textbf{Verdict: Violate}. \\ \cmidrule{2-3}
& \textbf{Defender} & The visual depicts a highly efficient learning method using a scale to emphasize the ``weight'' of knowledge. It is a rhetorical device for academic excellence and does not explicitly guarantee illegal admission outcomes. \textbf{Verdict: Comply}. \\ \cmidrule{2-3}
& \textbf{Umpire} & The conflict lies in whether ``Rapid Improvement'' constitutes inducement. Per \textbf{P33 (K12-T)}, the explicit use of ``Shortcut'' and ``Guaranteed Admission'' creates extreme anxiety and promises social mobility, meeting the definition of achievement-driven violation. \textbf{Verdict: Violate}. \\ \midrule
\multirow{5}{*}{\begin{minipage}{.18\textwidth} \centering \includegraphics[width=0.8\linewidth]{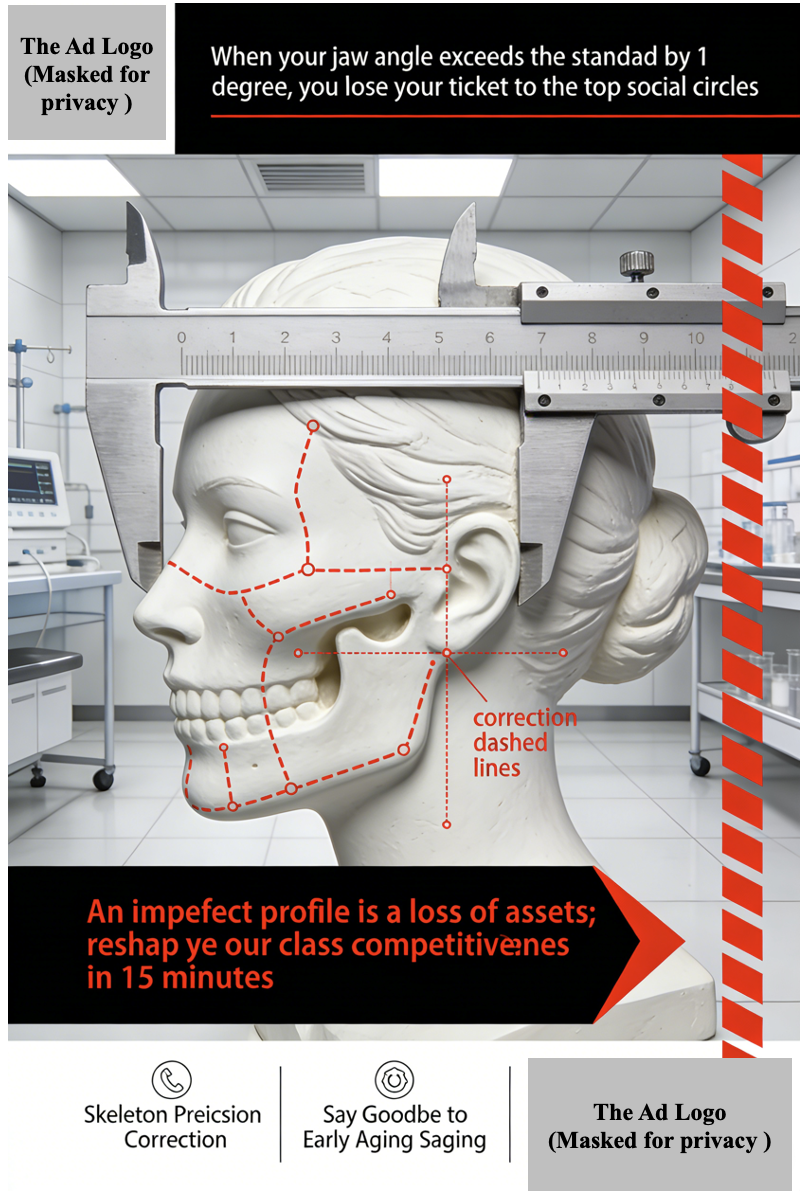} \\ \textbf{Case 2: Aest-A} \end{minipage}} 
& \textbf{Skeptic} & Although the ad uses a plaster model instead of a human, its logic that ``losing 1 degree means losing your ticket to top social circles'' is extreme. Does this precise quantification of the jawline fall under ``pathological aesthetic standards'' prohibited by policy? \\ \cline{2-3}
& \textbf{Prosecutor} & The ad uses calipers to measure the face and defines natural features as ``imperfect profiles'' and ``loss of assets''. It explicitly links appearance to social class, inducing self-loathing and urgency. \textbf{Verdict: Violate}. \\ \cmidrule{2-3}
& \textbf{Defender} & The visual is a professional simulation for medical education. Mentioning ``social circles'' is merely a rhetorical way to describe professional competitiveness and does not force a mandatory standard on all individuals. \textbf{Verdict: Comply}. \\ \cmidrule{2-3}
& \textbf{Umpire} & The Skeptic's point on ``quantified aesthetics'' is decisive. According to \textbf{P34 (Aest-A)}, inducing anxiety via quantified metrics is forbidden. Linking surgical reshaping to ``class competitiveness'' bypasses rational decision-making through psychological pressure. \textbf{Verdict: Violate}. \\ \bottomrule
\end{tabular}
\caption{Multi-agent dialectical reasoning analysis for representative violations.}
\label{tab_case}
\end{table*}

\section{Introduction}

Online advertising serves as the economic cornerstone of the modern internet, acting as a critical bridge between brands and global consumers. Given the proactive nature of algorithmic dissemination, maintaining a rigorous governance framework is essential for ensuring legal compliance and user safety. Unlike general content moderation \cite{gillespie2020content, gorwa2020algorithmic, langvardt2017regulating}, a single non-compliant advertisement can achieve massive reach through algorithmic pushing, necessitating a governance system that is not only precise but also highly responsive to an evolving regulatory landscape.

However, a fundamental challenge in ad governance is that policies are never static. Driven by seasonal trends and emerging social hotspots, new mandates, such as restrictions on K12 Achievement-Driven Tutoring  or Body \& Aesthetic Anxiety (as shown in Table \ref{tab:final_policies}), frequently emerge to address novel violations. Adapting existing models to these newly-added policies is a mission-critical yet formidable task. While platforms can collect sparse ``gold data'' reflecting new mandates, these samples are often insufficient to recalibrate large-scale models without compromising performance on historical data.

Designing such a policy-adaptive system entails three significant hurdles. 1) Label Inconsistency: historical data, though vast, was labeled under outdated policies ($\mathcal{P}_{old}$). Many samples may technically violate new dimensions ($\Delta \mathcal{P}$) but remain marked as compliant, leading to severe gradient conflicts. 2) Reasoning Ambiguity: new policies often involve subtle, ``gray-area'' interpretations where binary labels are insufficient for the model to internalize the underlying logic. 3) Recovery of Hard Samples: while models can identify overt violations, they often fail to recall sophisticated or deceptive non-compliant ads hidden within massive historical streams.

In this paper, we propose \textbf{ARGUS}, a Policy-Adaptive Ad Governance system that enables Evolving Reinforcement through Multi-Agent Adversarial Umpiring. To address label inconsistency, ARGUS employs a multi-stage alignment process. After an initial \textit{Policy Seeding} phase, we introduce a \textbf{Prosecutor-Defender-Umpire} architecture. For historical samples, ARGUS utilizes an adversarial Vision-Language Model (VLM)  to act as a defender, generating counter-arguments for compliance against the model's prosecution. A neutral Umpire VLM then adjudicates these conflicting Chains-of-Thought (CoT), incorporating RAG-enhanced policy knowledge to rectify labels and provide high-fidelity rewards for reinforcement learning (RL). Furthermore, to unearth latent violations, ARGUS identifies latent non-compliant candidates, samples exhibiting high probabilistic affinity for new policies despite compliant labels. By subjecting these candidates to a tripartite dialectical discussion, ARGUS iteratively refines its reward model, synchronizing its reasoning pathways with the evolving regulatory mandates. Table \ref{tab_case} shows two cases of  multi-agent dialectical reasoning analysis for representative violations.

Our contributions are: 1) We propose the ARGUS, which achieves autonomous policy adaptation by resolving label inconsistencies between historical data and newly-emerging multi-dimensional policies.  2)  We introduce an Adversarial Umpiring mechanism that utilizes multi-agent dialectical reasoning and CoT synthesis as a dynamic reward function for reinforcement learning. 3)  Extensive experiments on both industrial and public datasets demonstrate that ARGUS significantly outperforms traditional fine-tuning baselines, achieving superior policy-adaptive learning with minimal gold data.

\section{Related Work} \label{sec_related}

\subsection{Advertisement Governance}

Unlike general content moderation tasks such as hate speech or toxicity detection \cite{toxvidlm, wang2026multi,toxicn, toxicloakcn,lu2024towards,wang2024mllm,wang2026streamsense,lu2026decoding}, advertisement governance presents unique challenges due to the highly adversarial nature of malicious actors and the subtle semantic ambiguity of compliance rules. In recent years, a growing body of research has emerged to address these complexities. For instance, RAVEN \citep{raven} first introduced a robust reinforcement learning (RL) framework for temporal localization in video ads, utilizing a multi-stage training scheme that combines coarse and fine-grained annotations. Building upon this, RAVEN++ \citep{raven++} focused on finer granularity and proposed an active reinforcement learning approach to proactively identify high-value samples during training, thereby optimizing localization boundaries. Furthermore, Hi-Guard \citep{higuard} established a trustworthy multimodal moderation framework through hierarchical labeling and rule-based knowledge injection, while BLM-Guard \citep{blmguard} leveraged RL to score the model's reasoning process based on dynamic principles. However, all aforementioned works operate under the assumption of static governance rules. They do not explore a framework capable of adapting to significant policy shifts and evolving mandates. To the best of our knowledge, this paper is the first to propose an evolutionary multi-agent dialectic framework specifically designed to handle the continuous shift and expansion of advertising policies in an industrial setting.

\begin{figure*}[!ht]
      \centering
      \includegraphics[width=1\linewidth]{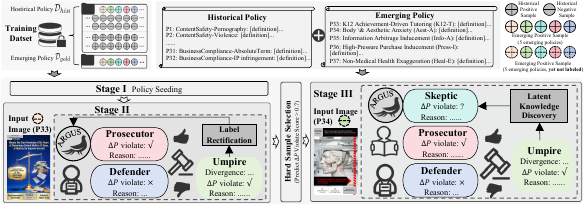}
      \caption{Overview of the ARGUS. ARGUS transitions through three stages: \textbf{Stage I (Policy Seeding)} for foundational policy perception; \textbf{Stage II (Adversarial Label Rectification)} which utilizes a bilateral debate between a \textbf{Prosecutor} and a \textbf{Defender} to rectify stale historical labels; and \textbf{Stage III (Latent Knowledge Discovery)} which employs a tripartite architecture including a \textbf{Skeptic} to unearth hard-to-detect violations in the uncertainty zone through logical triangulation by the \textbf{Umpire}.  Here, ``Positive'' indicates  non-compliant (violative), ``Negative'' indicates compliant (non-violative). The detailed case study is in Table \ref{tab_case} in the Appendix.}
      \label{fig_train_overview}
    \end{figure*}

\subsection{Reinforcement Learning in Large Models}

The landscape of Multimodal Large Language Models (MLLMs) \cite{yin2023survey, amini2024direct, chen2024tomgpt, zhang2024notellm2, ji2024tree,xu2024large, mixtral} has been fundamentally reshaped, and establishes the bedrock for cross-modal alignment and temporal reasoning \cite{yin2023survey, pentina2015curriculum,dlpl,llava, zhu2025llafs++,qwen_vl, zhu2024llafs, nyeshow, wei2022chain,zhuretrv}. Beyond general cross-modal understanding, complex visual perception scenarios demand fine-grained feature modeling and robust spatial reasoning capabilities \cite{soviany2022curriculum,popen,shazeer2017outrageously,skysenseo,sstkd_pami,graves2017automated,stlnet,mckinzie2024mm1,sstkd,hacohen2019power,ma2024fine,curriculum_rl,vicuna}. Such visual-oriented tasks present unique optimization challenges that vanilla pre-training fails to resolve, calling for dedicated alignment strategies to unify low-level visual sensing and high-level semantic comprehension \cite{urur,zhu2024ibd,gpwformer}. Reinforcement Learning (RL) has emerged as a critical paradigm for aligning model outputs with complex, high-dimensional human preferences \cite{yu2024rlhf, dpo, liustatistical}.

However, applying RL to the specific domain of advertisement governance introduces unique structural challenges that traditional RLHF frameworks \cite{rlhf} rarely encounter. First, semantic policy ambiguity creates a reward-shaping dilemma; unlike objective tasks such as coding or mathematics, advertising compliance relies on nuanced interpretations of ``harmfulness'' or ``deception'', leading to high-variance reward signals. 
Second, the adversarial evolution of violations necessitates that RL agents go beyond simple pattern recognition to develop robust counter-reasoning against intentional obfuscation. Third, the dynamic policy shift in industrial environments requires RL frameworks to maintain high semantic plasticity, integrating emerging regulations while strictly preserving historical governance logic.  ARGUS addresses these gaps by shifting from a monolithic reward model to a tri-party evolutionary game, where the reward is derived from a structured linguistic debate rather than a scalar preference score.

\section{Methodology}

\subsection{Problem Formulation}

We define the ad governance task as a multi-dimensional mapping function $f: \mathcal{X} \rightarrow (\mathbf{y}, \mathcal{C})$. Here, $\mathcal{X}$ is the space of multi-modal ads, and $\mathbf{y} = \{y^{(k)}\}_{k \in \mathcal{K}}$ is a vector of compliance labels, where $y^{(k)} \in \{0, 1\}$ indicates whether the ad violates the $k$-th specific policy category in the set $\mathcal{K}$. $\mathcal{C}$ denotes the corresponding Chain-of-Thought (CoT) reasoning that justifies the labels.

In a dynamic regulatory environment, the policy set expands from $\mathcal{P}_{old}$ to $\mathcal{P}_{new}$, where $\Delta \mathcal{P} = \{p_1, p_2, \dots, p_n\}$ represents $n$ newly-emerging policy mandates (e.g., K12-T, Aest-A, as shown in Table \ref{tab:final_policies}). This evolution implies that the label space $\mathcal{K}$ is non-stationary, effectively adding new dimensions to the compliance vector $\mathbf{y}$.

Given a massive historical dataset $\mathcal{D}_{hist} = \{(x_i, \mathbf{y}_i^{old})\}_{i=1}^N$ labeled under $\mathcal{P}_{old}$, and a sparse ``gold'' dataset $\mathcal{D}_{gold} = \{(x_j, \mathbf{y}_j^{new})\}_{j=1}^M$ ($M \ll N$) specifically annotated for the emerging categories in $\Delta \mathcal{P}$, the objective is to optimize a model $f_\theta$ to align with $\mathcal{P}_{new}$. The core challenge is label multi-dimensionality and inconsistency: a historical sample $x \in \mathcal{D}_{hist}$ marked as compliant under $\mathcal{P}_{old}$ may now possess a positive violation label $y^{(k)}=1$ for some $k \in \Delta \mathcal{P}$. The model must therefore learn to infer these latent labels across multiple new policy dimensions \textbf{without exhaustive manual re-annotation} of $\mathcal{D}_{hist}$.

\subsection{Evolving Reinforcement Framework}

To bridge the gap between static training and dynamic mandates, we propose \textbf{Evolving Reinforcement}, a three-stage framework powered by Group Relative Policy Optimization (GRPO) \cite{shao2024deepseekmath}. Unlike traditional RL training \cite{rlhf}, ARGUS \textit{evolves the policy and reward signals in tandem}, transitioning from initial perception to adversarial rectification and finally to latent discovery.

\textbf{Stage I: Policy Seeding.}  
This is a warm-up perception stage, we initialize the model via supervised alignment by blending $\mathcal{D}_{gold}$ with a curated subset of $\mathcal{D}_{hist}$. This phase establishes the foundational ``policy sense'' ($f_{\theta_{base}}$), ensuring the model internalizes the basic semantic boundaries of the new dimensions in $\Delta \mathcal{P}$ before entering high-variance reinforcement stages.

\textbf{Stage II: Adversarial Label Rectification.} 
This stage resolves \textit{label inconsistencies} in historical data through active reinforcement. For samples in $\mathcal{D}_{hist}$, the model acts as a Prosecutor to challenge outdated labels. A \textit{Prosecutor-Defender-Auditor} game generates dynamic rewards $R_{rect}$ by adjudicating multi-agent rationales. This process effectively ``overwrites'' stale historical noise with high-fidelity rewards aligned with $\mathcal{P}_{new}$.

\textbf{Stage III: Latent Knowledge Discovery.} 
The final stage targets \textit{hard samples} within the ``uncertainty zone'' (cases where violation intent is deeply latent or deceptive). By subjecting these candidates to a tripartite dialectic, the Auditor synthesizes conflicting views to refine decision boundaries in complex territories. This completes the evolution from surface-level pattern matching to deep, policy-driven reasoning.

\begin{figure*}[!ht]
      \centering
      \includegraphics[width=1\linewidth]{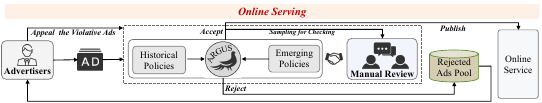}
      \caption{The online deployment of ARGUS.}
      \label{fig_deploy}
    \end{figure*}

\subsection{Stage I: Policy Seeding}
\label{subsec:stage1}

The goal of Stage I is \textit{Incremental Policy Calibration}: enabling the production-grade model $f_{\theta}$ (pre-aligned with $\mathcal{P}_{old}$) to perceive the boundaries of $\Delta \mathcal{P}$ without degrading historical performance. 

\textbf{Strategic Data Blending. }
To incorporate new knowledge while avoiding \textit{Gradient Overwhelming} from stale labels, we construct a hybrid training set $\mathcal{D}_{SFT} = \mathcal{D}_{gold} \cup \mathcal{D}'_{hist}$. Here, $\mathcal{D}'_{hist}$ is a representative subset ($\approx 40\%$) of historical data. This ratio ensures that the sparse but precise signal from $\mathcal{D}_{gold}$ is not drowned out by historical samples that might technically violate $\Delta \mathcal{P}$ but are marked as compliant in the legacy logs.

\textbf{Optimization Objective. }
We optimize $f_{\theta}$ to maximize the joint likelihood of multi-dimensional compliance labels $\mathbf{y}$ and reasoning paths $\mathcal{C}$ under the expanded policy $\mathcal{P}_{new}$:
\begin{equation}
    \mathcal{L}_{stage1}(\theta) = - \sum_{(x, \mathbf{y}, \mathcal{C}) \in \mathcal{D}_{SFT}} \log P(\mathbf{y}, \mathcal{C} | x, \mathcal{P}_{new}; \theta)
\end{equation}
This yields a seeded model $f_{\theta_{base}}$ that possesses a preliminary ``policy sense'' for the new categories in $\Delta \mathcal{P}$.

\textbf{Transition to Evolution. }
While Stage I enables the identification of overt violations, $f_{\theta_{base}}$ inevitably encounters \textbf{logical dissonance} where its emerging perception of $\Delta \mathcal{P}$ contradicts stale labels in $\mathcal{D}_{hist}$. This dissonance is not treated as noise but as the necessary impetus for the adversarial dialectic in Stage II.

\subsection{Stage II: Adversarial Label Rectification}

Following the seeding stage, we initiate Adversarial Label Rectification to resolve explicit conflicts where historical labels in $\mathcal{D}_{hist}$ contradict the emerging logic of $\Delta \mathcal{P}$.

\textbf{Dialectical Debate. } 
We establish a competitive reasoning environment to ``stress-test'' historical labels. The current policy model acts as the \textbf{Prosecutor}, generating a CoT that identifies violations within $\Delta \mathcal{P}$. Conversely, a strong VLM serves as the \textbf{Adversarial Defender}, tasked with justifying compliance through benign interpretations. This bilateral debate ensures that the model does not become over-sensitized to new polices while ignoring valid creative nuances.

\textbf{Umpire Adjudication. } 
The conflicting rationales are submitted to a neutral \textbf{Umpire VLM}. To ensure grounded decision-making, the Umpire utilizes a RAG-enhanced mechanism to retrieve specific clauses from $\Delta \mathcal{P}$ and reference samples from $\mathcal{D}_{gold}$. The Umpire adjudicates the debate by evaluating logical rigor and policy adherence:
\begin{equation}
(y^*, \mathcal{C}^*) = \text{Umpire}(CoT_{pros}, CoT_{def} \mid \mathcal{P}_{new})
\end{equation}
The output consists of a rectified label $y^*$ and a high-fidelity reasoning chain $\mathcal{C}^*$.

In this stage, we redefine the reward function to guide the policy away from historical bias. The reward is composed of two components: the \textit{Historical Consistency Reward} ($R_{hist}$) and the \textit{Dialectic Rectification Reward} ($R_{rect}$). For $R_{hist}$,  initially, the reward is derived from the legacy labels $y_{old} \in \mathcal{D}_{hist}$. However, as discussed, $R_{hist}$ contains "stale noise" that contradicts $\Delta \mathcal{P}$. For $R_{rect}$, the Umpire adjudicates the debate to produce a gold-standard tuple $(y^*, \mathcal{C}^*)$. The reward for a model-generated sample $(y, \mathcal{C})$ is then computed as:
    \begin{equation}
        R_{rect}(y, \mathcal{C}) = \mathbf{1}(y = y^*) + \text{sim}(\mathcal{C}, \mathcal{C}^*)
    \end{equation}
    where $\text{sim}(\cdot)$ measures the semantic alignment between the model's reasoning and the Umpire's adjudicated CoT.

By utilizing $R_{rect}$ as the primary optimization target in GRPO, $f_\theta$ is forced to resolve the ``logical dissonance'' encountered in Stage I. The model learns to prioritize the umpire’s adjudicated logic over the noisy $y_{old}$, achieving a high-quality self-corrected training stream. This ensures that both the final verdict $\mathbf{y}$ and the underlying reasoning $\mathcal{C}$ are synchronized with the emerging policy $\mathcal{P}_{new}$.

\subsection{Stage III: Latent Knowledge Discovery}

While Stage II resolves explicit label inconsistencies, sophisticated violations, such as deceptive creatives employing subtle evasion techniques, often persist in the decision boundary's ``gray area.'' In these cases, the model $f_\theta$ may exhibit a high probabilistic affinity toward $\Delta \mathcal{P}$ yet still output a compliant label due to cautious bias. Stage III aims to evolve the model's discriminative depth by unearthing these hard samples.

\textbf{Latent Candidate Selection.} 
To identify hard samples within the vast $\mathcal{D}_{hist}$, we leverage the model's internal confidence. We define latent non-compliant candidates as samples predicted as compliant ($y^{(k)}=0$) but possessing a posterior probability for an emerging policy $k \in \Delta \mathcal{P}$ that exceeds an empirical threshold $\tau$:
\begin{equation}
\begin{aligned}
\mathcal{D}_{latent} = \{ & x \in \mathcal{D}_{hist} \mid y^{(k)} = 0 \text{ and } \\ &P(y^{(k)}=1|x) > \tau\}.
\end{aligned}
\end{equation}
These candidates represent the ``uncertainty zone'' where the model's latent representation contradicts its categorical prediction.

\begin{table*}[t]
\centering
\scalebox{0.67}{\begin{tabular}{l|cc|cccccccccc|cc}
\toprule
\multirow{2}{*}{\textbf{Method}} & \multicolumn{2}{c|}{\textbf{Historical Overall}} & \multicolumn{10}{c|}{\textbf{Emerging Policies ($\Delta \mathcal{P}$)}} & \multicolumn{2}{c}{\textbf{Avg. $\Delta \mathcal{P}$}} \\
\cmidrule{2-15}
 & \textbf{Prec.} & \textbf{Rec.} & \multicolumn{2}{c}{\textbf{K12-T}} & \multicolumn{2}{c}{\textbf{Aest-A}} & \multicolumn{2}{c}{\textbf{ Info-A}} & \multicolumn{2}{c}{\textbf{Press-I}} & \multicolumn{2}{c|}{\textbf{ Heal-E}} & \textbf{Prec.} & \textbf{Rec.} \\
 & & & \textbf{Prec.} & \textbf{Rec.} & \textbf{Prec.} & \textbf{Rec.} & \textbf{Prec.} & \textbf{Rec.} & \textbf{Prec.} & \textbf{Rec.} & \textbf{Prec.} & \textbf{Rec.} & & \\ \midrule
\multicolumn{15}{l}{\textit{\textbf{Historical Expert (Qwen3-VL-8B)}}} \\
SFT (on $\mathcal{D}_{hist}$ only) & \textbf{0.842} & \textbf{0.858} & 0.352 & 0.421 & 0.312 & 0.385 & 0.344 & 0.412 & 0.451 & 0.512 & 0.412 & 0.485 & 0.374 & 0.443 \\ \midrule
\multicolumn{15}{l}{\textit{\textbf{Zero-shot (Large Models)}}} \\
GPT-4o & 0.485 & 0.612 & 0.421 & 0.582 & 0.384 & 0.521 & 0.442 & 0.594 & 0.521 & 0.655 & 0.481 & 0.615 & 0.450 & 0.593 \\
Gemini 2.0 Flash & 0.462 & 0.594 & 0.402 & 0.551 & 0.365 & 0.504 & 0.425 & 0.572 & 0.501 & 0.632 & 0.462 & 0.581 & 0.431 & 0.568 \\
Qwen2.5-VL-72B & 0.455 & 0.582 & 0.412 & 0.542 & 0.375 & 0.495 & 0.415 & 0.561 & 0.492 & 0.622 & 0.455 & 0.574 & 0.430 & 0.559 \\
Qwen3-235B-A22B & 0.512 & 0.635 & 0.454 & 0.612 & 0.412 & 0.554 & 0.478 & 0.642 & 0.572 & 0.704 & 0.521 & 0.641 & 0.487 & 0.631 \\ \midrule
\multicolumn{15}{l}{\textit{\textbf{Zero-shot (Base Models)}}} \\
Qwen2.5-VL-7B & 0.312 & 0.451 & 0.312 & 0.484 & 0.285 & 0.454 & 0.344 & 0.511 & 0.421 & 0.582 & 0.374 & 0.522 & 0.347 & 0.511 \\
Qwen3-VL-8B & 0.342 & 0.482 & 0.342 & 0.512 & 0.312 & 0.485 & 0.375 & 0.542 & 0.454 & 0.612 & 0.402 & 0.552 & 0.377 & 0.541 \\ \midrule
\multicolumn{15}{l}{\textit{\textbf{Fine-tuning (Qwen2.5-VL-7B)}}} \\
Vanilla SFT (on $\mathcal{D}_{gold}$) & 0.421 & 0.402$^\dagger$ & 0.762 & 0.745 & 0.714 & 0.682 & 0.735 & 0.712 & 0.782 & 0.754 & 0.751 & 0.732 & 0.749 & 0.725 \\
SFT + Replay (40\%) & 0.772 & 0.765 & 0.735 & 0.722 & 0.692 & 0.665 & 0.715 & 0.692 & 0.782 & 0.751 & 0.741 & 0.725 & 0.733 & 0.711 \\
EWC (Continual Learning) & 0.792 & 0.784 & 0.744 & 0.732 & 0.702 & 0.682 & 0.724 & 0.705 & 0.792 & 0.765 & 0.752 & 0.741 & 0.743 & 0.725 \\ \midrule
\multicolumn{15}{l}{\textit{\textbf{Fine-tuning (Qwen3-VL-8B)}}} \\
Vanilla SFT (on $\mathcal{D}_{gold}$) & 0.454 & 0.432$^\dagger$ & 0.782 & 0.761 & 0.735 & 0.702 & 0.754 & 0.731 & 0.815 & 0.782 & 0.784 & 0.764 & 0.774 & 0.748 \\
SFT + Replay (40\%) & 0.791 & 0.785 & 0.752 & 0.745 & 0.708 & 0.685 & 0.735 & 0.718 & 0.805 & 0.775 & 0.765 & 0.741 & 0.753 & 0.733 \\
EWC (Continual Learning) & 0.802 & 0.794 & 0.761 & 0.746 & 0.715 & 0.698 & 0.742 & 0.725 & 0.811 & 0.782 & 0.772 & 0.754 & 0.760 & 0.741 \\ \midrule
\textbf{ARGUS} (Qwen2.5-VL-7B) & 0.815 & 0.822 & 0.785 & 0.804 & 0.702 & 0.765 & 0.741 & 0.792 & 0.801 & 0.854 & 0.784 & 0.812 & 0.763 & 0.805 \\
\textbf{ARGUS} (Qwen3-VL-8B) & 0.828 & 0.841 & \textbf{0.812} & \textbf{0.835} & \textbf{0.734} & \textbf{0.792} & \textbf{0.782} & \textbf{0.824} & \textbf{0.834} & \textbf{0.885} & \textbf{0.815} & \textbf{0.842} & \textbf{0.795} & \textbf{0.836} \\ \bottomrule
\end{tabular}}
\caption{Results on Industrial Dataset. Prec. and Rec. denote Precision and Recall. Zero-shot models exhibit low precision and higher recall on $\Delta \mathcal{P}$ due to unfamiliarity with specific boundary nuances. $^\dagger$ indicates catastrophic forgetting, vanilla SFT shows strong adaptation to new rules but suffers from severe historical knowledge collapse. ARGUS-8B maintains a minimal performance gap ($<2\%$) compared to the Historical Expert while outperforming all baselines on average $\Delta \mathcal{P}$.}
\label{tab:main_results}
\end{table*}

\textbf{Tripartite Dialectical Reasoning.} 
To resolve high-entropy cases, we upgrade the adversarial mechanism, by introducing the current model $f_\theta$ as the \textbf{Skeptic}, which provides a ``doubt-based'' CoT that articulates the suspicious features triggering its uncertainty. This is combined with polarized perspectives from the agents introduced in Stage II, 
 \textbf{1) Prosecutor Agent:} Constructs a rigorous argument for violation based on the latest policy $\Delta \mathcal{P}$;  \textbf{2) Defender Agent:} Provides benign interpretations of the ad to prevent over-sensitization; \textbf{3) Skeptic ($f_\theta$):} Highlighting internal conflicts and the reasoning behind its high-probability latent state.

\textbf{Umpire Synthesis and RL Evolution.} 
The neutral Umpire VLM gathers the dialectical triplet $\{CoT_{pros}, CoT_{def}, CoT_{skeptic}\}$. Through logical triangulation, the Umpire synthesizes a rectified conclusion $y^*$ and a standardized reasoning chain $\mathcal{C}^*$. We define the \textbf{Evolution Reward} $R_{evolve}$ to align the model with this refined logic:
\begin{equation}
R_{latent}(y, \mathcal{C}) = \mathbf{1}(y = y^*) + \text{sim}(\mathcal{C}, \mathcal{C}^*)
\end{equation}
By optimizing against $R_{latent}$ via GRPO, the model's decision boundary is pushed into hard-negative territories. This completes the policy evolution from surface-level perception to deep, intent-driven policy deduction.

\section{Online Deployment}

Fig. \ref{fig_deploy} shows the online deployment of ARGUS, we include the detailed illustration in Sec. \ref{sec_deploy} in Appendix.

\section{Experiments}

\subsection{Offline Testing on Industrial Datasets}

As illustrated in Table~\ref{tab:main_results}, a comprehensive comparison between ARGUS and diverse baselines yields the following insights:

\textbf{Mitigating Catastrophic Forgetting.}
Maintaining stability on historical mandates during policy adaptation is a core challenge. The Vanilla SFT baseline exhibits severe knowledge erosion, with historical recall collapsing to $0.432$ ($^\dagger$). In contrast, ARGUS-8B retains a historical recall of $0.841$, matching the specialized Historical Expert ($0.858$) within a marginal $1.7\%$ gap. This confirms that our evolutionary framework effectively preserves foundational governance logic while incorporating new mandates.

\textbf{Domain Expertise vs. Model Scale.}
General-purpose giants like GPT-4o and Qwen3-235B exhibit a ``low-precision, high-recall'' bias, failing to outperform the specialized ARGUS-8B on emerging policies ($\Delta \mathcal{P}$). ARGUS achieves a precision improvement of over $30\%$ compared to these models, demonstrating that targeted adversarial evolution provides finer boundary sensitivity for industrial compliance than raw parameter scaling.

\textbf{Effectiveness of Adversarial Dialectic.}
While continual learning methods like EWC effectively mitigate forgetting, their adaptation to new policies is suboptimal. ARGUS-8B outperforms EWC \cite{ewc} in average $\Delta \mathcal{P}$ recall by $9.5\%$ ($0.836$ vs. $0.741$), with the most significant gains in high-ambiguity domains. This confirms that the tri-party game, specifically the Prosecutor's ability to synthesize ``gray-area'' cases, forces the model to internalize intent-level non-compliance rather than surface-level pattern matching.

\begin{table}[t]
\centering
\scalebox{0.63}{
\begin{tabular}{l|cc|cc}
\toprule
\multirow{2}{*}{\textbf{Method}} & \multicolumn{2}{c|}{\textbf{Historical Overall}} & \multicolumn{2}{c}{\begin{tabular}[l]{@{}l@{}}\textbf{Emerging: Dispirited } \\ \textbf{Culture (Avg. $\Delta \mathcal{P}$)}\end{tabular}} \\
\cmidrule{2-5}
 & \textbf{Prec.} & \textbf{Rec.} & \textbf{Prec.} & \textbf{Rec.} \\ \midrule
\multicolumn{5}{l}{\textit{\textbf{Historical Expert (Qwen3-VL-8B)}}} \\
SFT (on $\mathcal{D}_{hist}$ only) & \textbf{0.634} & \textbf{0.658} & 0.124 & 0.152 \\ \midrule

\multicolumn{5}{l}{\textit{\textbf{Zero-shot (Large Models)}}} \\
GPT-4o & 0.531 & 0.557 & 0.311 & 0.365 \\
Qwen3-VL-8B & 0.304 & 0.402 & 0.221 & 0.262 \\ \midrule

\multicolumn{5}{l}{\textit{\textbf{Fine-tuning (Qwen3-VL-8B)}}} \\
Vanilla SFT (on $\mathcal{D}_{gold}$) & 0.332 & 0.384$^\dagger$ & 0.415 & 0.435 \\
SFT + Replay (40\%) & 0.492 & 0.512 & 0.395 & 0.417 \\
EWC (Continual Learning) & 0.512 & 0.525 & 0.410 & 0.413 \\ \midrule

\textbf{ARGUS (Qwen3-VL-8B)} & 0.621 & 0.655 & \textbf{0.454} & \textbf{0.482} \\ \bottomrule
\end{tabular}
}
\caption{Results on the public ToxiCN MM dataset. ``Dispirited Culture'' serves as the emerging policy to test the model's adaptation to subtle linguistic metaphors. $^\dagger$ indicates catastrophic forgetting where vanilla SFT fails to retain historical knowledge. ARGUS maintains stability on historical domains while achieving best results on emerging toxic memes.}
\label{tab:public_results}
\end{table}

\textbf{Foundation Model Impact.}
Comparing ARGUS variants, the Qwen3-8B backbone consistently yields a $2\%$--$3\%$ improvement over the Qwen2.5-7B version. While the framework is model-agnostic, a stronger visual-linguistic foundation provides superior ``logical intuition'', facilitating a more sophisticated equilibrium during the policy evolution process.

\subsection{Offline Testing on Public Datasets}

To validate cross-domain generalization, we evaluated ARGUS on the public ToxiCN MM dataset. Results in Table~\ref{tab:public_results} mirror industrial trends. On the one hand, the \textit{Historical Expert} fails on the emerging \textit{Dispirited Culture} policy (0.152 recall), as standard safety models are ``blind'' to subtle cultural metaphors. 
\begin{table}[h]
\centering
\scalebox{0.8}{
\begin{tabular}{lccc}
\toprule
\textbf{Online A/B Testing Group} & \textbf{VLR} $\downarrow$ & \textbf{AAR} $\uparrow$ & \textbf{FPR} $\downarrow$ \\ \midrule
Control Group (Production) & 1.42\% & 68.5\% & 0.35\% \\
Treatment Group (\textbf{ARGUS}) & 0.92\% & 76.2\% & 0.32\% \\ \midrule
Relative Improvement & +35.2\% & +11.2\% & +8.5\% \\ \bottomrule
\end{tabular}
}
\caption{Online A/B testing results. ARGUS significantly reduces violation leakage while improving audit automation.}
\label{tab:ab_test}
\end{table}
Conversely, while \textit{Vanilla SFT} adapts to the new category, it suffers from catastrophic forgetting, with historical recall plunging to 0.384. ARGUS achieves a superior balance, maintaining near-optimal historical recall (0.655) while effectively integrating new policy knowledge. On the other hand, general models like GPT-4o lack the policy-sensitivity required for specialized Chinese memes, with ARGUS outperforming GPT-4o by 11.7\% in recall on the emerging category. Compared to EWC  and SFT+Replay, ARGUS demonstrates higher plasticity; its 6.9\% absolute improvement over EWC highlights that the tri-party game is not mere regularization, but a reasoning-driven adaptation process that resolves logical contradictions between evolving mandates.

\subsection{Online A/B Testing}

To evaluate the practical utility of ARGUS, we conduct the online A/B experiment on our advertising platform. We allocate a small amount of the production traffic to the Treatment group, while the remaining traffic is handled by the Control group. Both the groups 
are based on Qwen3-VL 8B model. We evaluated three primary industrial metrics: (1) Violation Leakage Rate (VLR), determined by human expert back-checking; (2) Audit Automation Rate (AAR), the ratio of ads processed without human intervention; and (3) False Positive Rate (FPR), verified by professional auditors. As shown in Table~\ref{tab:ab_test}, ARGUS achieves a 35.2\% reduction in VLR, and the 11.2\% increase in AAR significantly reduced human labor costs by moving high-confidence ``gray-area'' traffic into automated pipelines. Crucially, these gains are achieved alongside a lower FPR (0.32\%) as verified by auditors, ensuring that the platform’s efficiency is improved without compromising the experience of compliant advertisers.

\section{Conclusion}

In this paper, we present ARGUS, an evolutionary framework designed for ad governance under non-stationary policy environments. ARGUS employs a multi-agent tri-party dialectic to transform passive policy compliance into an active process. Extensive evaluations on both industrial and public datasets demonstrate that ARGUS achieves superior adaptation to emerging policies.

\section{Ethical Considerations}

Our research strictly adheres to established ethical guidelines and prioritizes data integrity and user privacy. The datasets utilized in this study consist of publicly accessible or de-identified industrial advertising samples. All labeling and evaluation processes are conducted solely for scientific analysis and do not reflect the institutional stances or personal opinions of the authors. Furthermore, the ARGUS framework is designed to enhance digital safety and platform reliability; however, we emphasize that its deployment should remain under human-in-the-loop oversight to prevent unintended biases. All resources and methodologies described herein are intended for research purposes, contributing to the broader goal of fostering a secure, compliant, and transparent digital advertising ecosystem.

\section*{Acknowledgment}
This research was supported in part by the National Natural Science Foundation of China (Nos. 625B2033). We acknowledge Baoyan Zhuang and Pengda Qin from Tencent for collaborating on data resources and application scenarios to validate and improve
algorithm performance.

\bibliography{custom}

\newpage

\appendix

\section*{Appendix}

\section{Online Deployment} \label{sec_deploy}

As shown in Fig. \ref{fig_deploy}, the online infrastructure employs a cascaded filtering mechanism to maintain low-latency response:  1) Initial Screening: Incoming ads from advertisers undergo a initial check. Ads triggered by this process are immediately rejected and returned to the advertiser. 2) ARGUS Adaptive Governance: Ads passing the initial filter are processed by the ARGUS engine, which specifically evaluates compliance against both Historical and Emerging Policies ($\Delta \mathcal{P}$). By leveraging the reasoning pathways evolved during offline stages, the engine can accurately identify novel violations that legacy systems might miss. 3) Manual Review \& Feedback: To ensure the highest fidelity, a specialized Manual Review module performs sampling for checking. This human-in-the-loop component serves as a final quality gate, and the resulting high-quality labels are fed back to the offline modeling pipeline to facilitate continuous model evolution. 4 Final Decision: Ads verified as compliant are published to the Online Service, while non-compliant samples are diverted to the Rejected Ads Pool.

\section{Datasets}

\subsection{Industrial Dataset}

To evaluate ARGUS in a rigorous production-level environment, we construct a large-scale multimodal (image \& text)  dataset derived from our advertising platform. The dataset is specifically designed to reflect the non-stationary nature of advertising governance, categorizing policies into two primary sets: \textbf{Historical Policies} ($\mathcal{P}_{hist}$) and \textbf{Emerging Policies} ($\Delta \mathcal{P}$).

\subsubsection{Emerging Policy Definitions ($\Delta \mathcal{P}$)}

To evaluate ARGUS's adaptability to non-stationary environments, we define five emerging policy domains ($\Delta \mathcal{P}$) that represent recent regulatory shifts and sophisticated deceptive tactics. These categories are characterized by high semantic ambiguity and require deep intent-level reasoning, as shown in Table \ref{tab:final_policies}.

\subsubsection{Data Composition and Statistics}
The dataset comprises approximately $680,000$ high-quality samples, as detailed in Table~\ref{tab:data_stats}: 
\begin{itemize}
    \item \textbf{Historical Policies ($\mathcal{P}_{hist}$):} This subset contains $168,000$ positive samples covering over 50 foundational categories. These include \textit{Basic Content Safety} (e.g., pornography, violence, bloodiness, hate speech, and prohibited items) and \textit{Business Compliance} (e.g., absolute terminology, exaggerated marketing, and IP infringement).
    \item \textbf{Emerging Policies ($\Delta \mathcal{P}$):} We collect $32,700$ samples representing five critical emerging policy shifts. These domains are characterized by their high semantic ambiguity and evolving deceptive tactics.
    \item \textbf{Negative Samples:} To ensure robust training and minimize false positives, we incorporated approximately $480,000$ negative samples (compliant ads) that share similar visual or linguistic features with non-compliant ones, creating a challenging "needle-in-a-haystack" scenario.
\end{itemize}

\begin{table*}[t]
\centering
\small
\begin{tabular}{lp{3.5cm}p{9.8cm}}
\toprule
\textbf{PolicyID} & \textbf{Policy Category} & \textbf{Description ($\Delta \mathcal{P}$)} \\ \midrule
\textbf{P33} & \textbf{K12 Achievement-Driven Tutoring (K12-T)} & Targets achievement-driven academic tutoring for K-12 students. It prohibits promoting "exam shortcuts" or "guaranteed admission" that exploit parental anxiety and utilitarian educational goals. \\ \midrule
\textbf{P34} & \textbf{Body \& Aesthetic Anxiety (Aest-A)} & Regulates content that promotes singular beauty standards (e.g., extreme thinness) or implies that physical flaws are barriers to a successful life, thereby inducing psychological distress and body dysmorphia. \\ \midrule
\textbf{ P35} & \textbf{Information Arbitrage Inducement (Info-A)} & Prohibits inducing financial investment through claims of "insider info" or "unclosed trends." It targets the masking of fraudulent risks under the guise of "wealth shortcuts" or exclusive "circle privileges." \\ \midrule
\textbf{P36 } & \textbf{High-Pressure Purchase Inducement (Press-I)} & Regulates the use of artificial urgency (e.g., fake countdowns, false stock limits) and compulsive logic (e.g., "regret for life") designed to bypass rational decision-making in e-commerce.\\ \midrule
\textbf{ P37} & \textbf{Non-Medical Health Exaggeration (Heal-E)} & Targets non-medical supplements claiming therapeutic effects (e.g., "curing cancer" or "restoring physiological indicators"). It prohibits substituting professional medical treatment with vague health-related efficacy claims. \\ \bottomrule
\end{tabular}
\caption{Definitions of the 5 emerging policies ($\Delta \mathcal{P}$). These categories represent high-stakes domains characterized by significant semantic ambiguity and adversarial evolution. The policy ids follow by the historical policies (total 32).}
\label{tab:final_policies}
\end{table*}

\begin{table}[h]
\centering
\scalebox{0.8}{
\begin{tabular}{lrc}
\toprule
\textbf{Category} & \textbf{Count} & \textbf{Type} \\ \midrule
Historical Policy Overall ($\mathcal{P}_{hist}$) & 168,000 & Positive \\ 
\midrule
\textit{Emerging Domains ($\Delta \mathcal{P}$)} & & \\
P33: K12-T & 4,500 & Positive \\
P34: Aest-A & 9,800 & Positive \\
P35: Info-A & 7,500 & Positive \\
P36: Press-I & 5,200 & Positive \\
P37: Heal-E & 5,700 & Positive \\
\midrule
Negative Samples (Compliant Ads) & 480,000 & Negative \\ \midrule
\textbf{Total} & \textbf{680,700} & -- \\ \bottomrule
\end{tabular}
}
\caption{Statistics of the industrial advertising dataset. Emerging policies focus on domains with high regulatory shifts and semantic ambiguity, while negative samples provide the necessary balance for robust industrial training.}
\label{tab:data_stats}
\end{table}

\begin{table}[h]
\centering
\scalebox{0.6}{
\begin{tabular}{lccc}
\toprule
\textbf{Evolving Reinforcement Stages} & \textbf{Hist. Rec.} & \textbf{\begin{tabular}[c]{@{}c@{}}Avg. $\Delta \mathcal{P}$ \\ Prec.\end{tabular}} & \textbf{\begin{tabular}[c]{@{}c@{}}Avg. $\Delta \mathcal{P}$ \\ Rec.\end{tabular}} \\ \midrule
Stage I: Policy Seeding & 0.785 & 0.753 & 0.733 \\
+ Stage II: (Adversarial Label Rectification) & 0.824 & 0.758 & 0.792 \\
+ Stage III: (Latent Knowledge Discovery) & \textbf{0.841} & \textbf{0.795} & \textbf{0.836} \\ 
\bottomrule
\end{tabular}}
\caption{Ablation study on the evolving reinforcement stages.}
\label{tab:ablation_stages}
\end{table}

\begin{table}[h]
\centering
\scalebox{0.85}{
\begin{tabular}{lcc}
\toprule
\textbf{Component Variant} & \textbf{Prec.} & \textbf{Rec.} \\ \midrule
Full ARGUS Dialectic & \textbf{0.795} & \textbf{0.836} \\
\textit{w/o} Prosecutor  & 0.732 & 0.695 \\
\textit{w/o} Defender  & 0.684 & 0.812 \\
\textit{w/o} Rationale (Labels Only) & 0.715 & 0.742 \\ \bottomrule
\end{tabular}
}
\caption{Component-wise ablation of the Multi-Agent Dialectic. Results are averaged across all $\Delta \mathcal{P}$ categories.}
\label{tab:component_ablation}
\end{table}

\begin{table}[h]
\centering
\scalebox{0.8}{
\begin{tabular}{lccc}
\toprule
\textbf{Adversarial Strategy} & \textbf{Std. SFT} & \textbf{GPT-4o} & \textbf{ARGUS} \\ \midrule
Normal Samples & 0.711 & 0.582 & \textbf{0.835} \\ \midrule
Adversarial Samples & 0.440 & 0.473 & \textbf{0.783} \\ \bottomrule
\end{tabular}
}
\caption{Detection recall under adversarial evasion.}
\label{tab:robustness}
\end{table}

\subsection{Public Dataset}

To evaluate the generalization and robustness of ARGUS on diverse social media content, we incorporate the ToxiCN MM dataset \citep{toxicn}, the first comprehensive Chinese harmful meme dataset. ToxiCN MM consists of 12,000 multimodal samples with fine-grained annotations across multiple dimensions of harmfulness on the Chinese internet. 

In our experimental setup, we adapt this dataset to simulate a policy evolution scenario by re-categorizing its original taxonomy into historical and emerging domains:
\begin{itemize}
    \item \textbf{Historical Policies ($\mathcal{P}_{hist}$):} We select 3 categories (\textit{Targeted Harmful}, \textit{General Offensive}, and \textit{Sexual Innuendo}) to represent the established historical policies. These categories align with the foundational content safety mandates in our industrial dataset.
    \item \textbf{Emerging Policy ($\Delta \mathcal{P}$):} We define \textit{Dispirited Culture} (Sang Culture) as the emerging policy. This category is particularly challenging as it involves subtle linguistic metaphors and pessimistic sentiments that deviate from traditional toxic content, requiring advanced reasoning to identify its potential negative social impact.
\end{itemize}

By bridging industrial advertising data with this public social media dataset, we create a rigorous benchmark that tests whether ARGUS can transfer its tri-party evolutionary logic from structured commercial policies to highly informal, culturally-specific internet memes. This dual-dataset evaluation ensures that the performance gains of ARGUS are not confined to a specific data distribution but are representative of a generalized capability in policy-driven content moderation.

\section{Dialectical Prompting Design} \label{sec_Prompting}
The efficacy of the adversarial evolution process hinges on the quality of the perspectives generated by the agents. We design a dialectical prompting scheme that enforces role-specific constraints, ensuring a high-entropy debate that covers the full spectrum of policy interpretation.

\textbf{Prosecutor: Rigorous Scrutiny.} 
The Prosecutor agent is prompted to act as a strict regulatory inspector. Its objective is to identify any potential violation of the newly emerging policy $\Delta \mathcal{P}$, no matter how subtle. We employ \textit{negative-bias prompting}, instructing the model to focus on deceptive visual layouts, exaggerated textual claims, and potential legal risks. 

\textbf{Defender: Adversarial Justification.} 
To prevent the system from becoming over-sensitive (i.e., excessive False Positive Rate), the Defender acts as ``sophisticated legal counsel''. It is tasked with providing alternative, benign interpretations for every point of contention raised by the Prosecutor. The prompt encourages \textit{Contextual Re-framing}, for instance, interpreting an exaggerated claim as ``artistic hyperbole'' or a high-pressure countdown as ``legitimate seasonal promotion''. By exploring the boundary of creative integrity, the Defender forces the debate to remain grounded, sharpening the model's ability to distinguish between ``gray-area'' creatives and true policy violations.

\textbf{Umpire: Logical Triangulation.} 
The Umpire is designed as a neutral adjudicator with \textit{RAG-enhanced objectivity}. Unlike the biased Prosecutor and Defender, the Umpire's prompt enforces a ``Fact-first, Logic-second'' hierarchy. It is required to first validate the specific policy clauses fetched from $\mathcal{P}_{new}$ and then evaluate the dialectical consistency of the opposing CoTs. We incorporate \textit{Reasoning Pruning} constraints, instructing the Umpire to explicitly reject any hallucinations or irrelevant arguments presented during the debate. The final output is a synthesized reasoning chain $\mathcal{C}^*$ that represents the optimal balance between regulatory rigor and creative tolerance.

\section{Ablation Study} \label{sec_abl}

\subsection{ Study on Evolving Reinforcement Stages}

To investigate the individual contribution of each stage within the ARGUS framework, we conduct an incremental ablation study. The results, summarized in Table~\ref{tab:ablation_stages}, demonstrate the cumulative performance gains from our multi-stage evolutionary strategy.

\textbf{Foundational Alignment via Stage I.}
Stage I (Policy Seeding) establishes the initial cross-domain alignment. It provides a baseline historical recall of 0.785 and an average $\Delta \mathcal{P}$ recall of 0.733. This stage ensures that the auditor agent internalizes the basic semantic boundaries of new policies before engaging in complex adversarial reasoning, preventing the model from starting the reinforcement process with unstable or biased gradients.

\textbf{Adversarial Rectification in Stage II.}
The introduction of the Stage II (Adversarial Label Rectification)  triggers a significant performance leap. These gains suggest that the dialectical conflict between the Prosecutor and Defender effectively rectifies mislabeled ``gray-area'' samples, forcing the model to develop a more robust reasoning logic rather than relying on surface-level pattern matching.

\textbf{Boundary Refinement via Stage III.}
Stage III (Latent Knowledge Discovery) further polishes the model's sensitivity by mining and synthesizing hard adversarial cases. This stage achieves the peak performance across all metrics. Notably, the continued improvement in Historical Recall  indicates that the hard sample mining process doesn't just aid new policy adaptation, it also reinforces the model's overall logical intuition, making it more resilient across the entire policy spectrum.

\subsection{Ablation on Multi-Agent Dialectic Components}

We conduct a component-wise ablation to verify the synergy between the agents. Results in Table~\ref{tab:component_ablation} highlight the distinct contribution of each role to the overall policy evolution.

\textbf{Precision-Recall Balance. } 
The Prosecutor and Defender serve as the primary drivers for Recall and Precision, respectively. Removing the Prosecutor leads to a 14.1\% drop in recall (0.695), as the model fails to uncover latent non-compliance. Conversely, removing the Defender causes precision to plunge to 0.684. This confirms that the Defender prevents ``over-censorship'' by forcing the model to consider benign justifications in gray-area cases.

\textbf{The Power of Rationale. } 
A crucial finding is that the \textbf{linguistic debate} itself is indispensable. When the agents provide only binary labels without detailed rationales (\textit{w/o Rationale}), performance drops across all metrics. This proves that ARGUS's strength lies in ``policy reasoning'' rather than simple pattern matching, the model requires the logical depth of the debate to internalize complex regulatory boundaries.

\subsection{Robustness against Adversarial Evasion}

In real-world scenarios, evasion techniques like homophone replacement and visual blurring are usually used to bypass the filters. We evaluate ARGUS on an Adversarial Evaluation Set (2,000 samples) featuring these sophisticated obfuscation strategies.
As shown in Table~\ref{tab:robustness}, ARGUS exhibits superior resilience compared to baselines. While the standard SFT model's recall plunges by 38.1\% (from 0.711 to 0.440) when facing adversarial samples, ARGUS maintains a robust average recall of 0.783, experiencing only a marginal 6.2\% decrease. This suggests that the tri-party game forces the model to focus on intent-level cues rather than surface-level patterns.

\section{Limitations and Future Work}

A primary constraint of this work is that ARGUS is currently evaluated on image-text advertisements. While effective for spatial reasoning, this focus does not account for the unique challenges of the video domain, which has become a dominant medium in digital marketing. Video governance requires capturing complex temporal dynamics and multi-modal synchronization, where non-compliance often emerges from the sequence of frames and audio cues rather than a single  image. Extending ARGUS to support temporal-aware reasoning and video-based policy evolution remains a key direction for our future research.

\end{document}